\documentclass[letterpaper, 10 pt, conference]{ieeeconf}  

\IEEEoverridecommandlockouts                              

\usepackage{url}
\usepackage{graphics} 
\usepackage{epsfig} 
\usepackage{mathptmx} 
\usepackage{times} 
\usepackage{amsmath} 
\usepackage{amssymb}  
\usepackage{subfigure}
\usepackage[T1]{fontenc}
\usepackage[vlined,ruled]{algorithm2e}
\usepackage{booktabs}
\usepackage{multirow}

\usepackage{graphicx}
\usepackage{float}

\usepackage{cite}
\usepackage{xcolor}
\usepackage{url}
\usepackage{algpseudocode}
\usepackage[colorlinks,linkcolor=blue,anchorcolor=blue,citecolor=green]{hyperref} 
\usepackage[export]{adjustbox}
\usepackage{balance}

\title{\LARGE \bf
Event-VPR: End-to-End Weakly Supervised Network Architecture for Event-based Visual Place Recognition
}

\author{Delei Kong$^{1}$, Zheng Fang$^{2}$, Haojia Li$^{2}$, Kuanxu Hou$^{2}$, Sonya Coleman$^{3}$ and Dermot Kerr$^{3}$
\thanks{*This work was supported by National Natural Science Foundation of China (No. 62073066), Science and Technology on Near-Surface Detection Laboratory (No. 6142414200208), the Fundamental Research Funds for the Central Universities(No.N182608003, No. N172608005).}
\thanks{$^{1}$Delei Kong is with College of Information Science and Engineering, Northeastern University, Shenyang, China. }%
\thanks{$^{2}$Zheng Fang, Haojia Li and Kuanxu Hou are with Faculty of Robot Science and Engineering, Northeastern University, Shenyang, China. Corresponding author: {\tt\small fangzheng@mail.neu.edu.cn}}
\thanks{$^{3}$Sonya Coleman and Dermot Kerr are with Faculty of Computing, Engineering and Built Environment, Ulster University, Northern Ireland, UK. }%
}

\begin{document}

\maketitle
\thispagestyle{empty}
\pagestyle{empty}

\begin{abstract}
Traditional visual place recognition (VPR) methods generally use frame-based cameras, which is easy to fail due to dramatic illumination changes or fast motions. In this paper, we propose an end-to-end visual place recognition network for event cameras, which can achieve good place recognition performance in challenging environments. The key idea of the proposed algorithm is firstly to characterize the event streams with the EST voxel grid, then extract features using a convolution network, and finally aggregate features using an improved VLAD network to realize end-to-end visual place recognition using event streams. To verify the effectiveness of the proposed algorithm, we compare the proposed method with classical VPR methods on the event-based driving datasets (MVSEC, DDD17) and the synthetic datasets (Oxford RobotCar). Experimental results show that the proposed method can achieve much better performance in challenging scenarios. To our knowledge, this is the first end-to-end event-based VPR method. The accompanying source code is available at \url{https://github.com/kongdelei/Event-VPR}.
\end{abstract}

\section{INTRODUCTION}
\label{sec:introduction}
Visual place recognition (VPR) \cite{lowry2015visual} is a very challenging problem in the field of computer vision and mobile robots. In computer vision, VPR can be used to retrieve visual information or cross-time place information in a large-scale image database with geographic information annotation, or be used in interactive 3D vision applications such as augmented reality (AR). In mobile robots, the ability of robots to recognize visual places in GPS-denied environment is one of the key capabilities for autonomous localization and navigation. In simultaneous localization and mapping (SLAM), VPR is an important component of loop closure detection \cite{angeli2008fast} \cite{galvez2012bags}, which can be used to detect candidate loop-closures and eliminate accumulated errors through global optimization for globally consistent pose estimating and mapping.

At present, there are many solutions to the VPR problem in large-scale environments. For those existing solutions, monocular, binocular, panorama camera and other frame-based vision sensors are widely used. However, since frame-based vision sensors suffer from issues such as illumination changes, motion blur and redundant information, which makes traditional VPR methods difficult to deal with recognition tasks in some challenging environments. Besides, most existing methods are based on the appearance of scenes \cite{oertel2020augmenting}. Due to various reasons, such as day-and-night, weather and seasonal changes, the appearance of the same place will change greatly at different time. In addition, the appearance of some places at long distances may be very similar. These issues pose great challenges to the existing large-scale frame-based VPR methods.

\begin{figure}[t]
\centering
\includegraphics[width=\columnwidth]{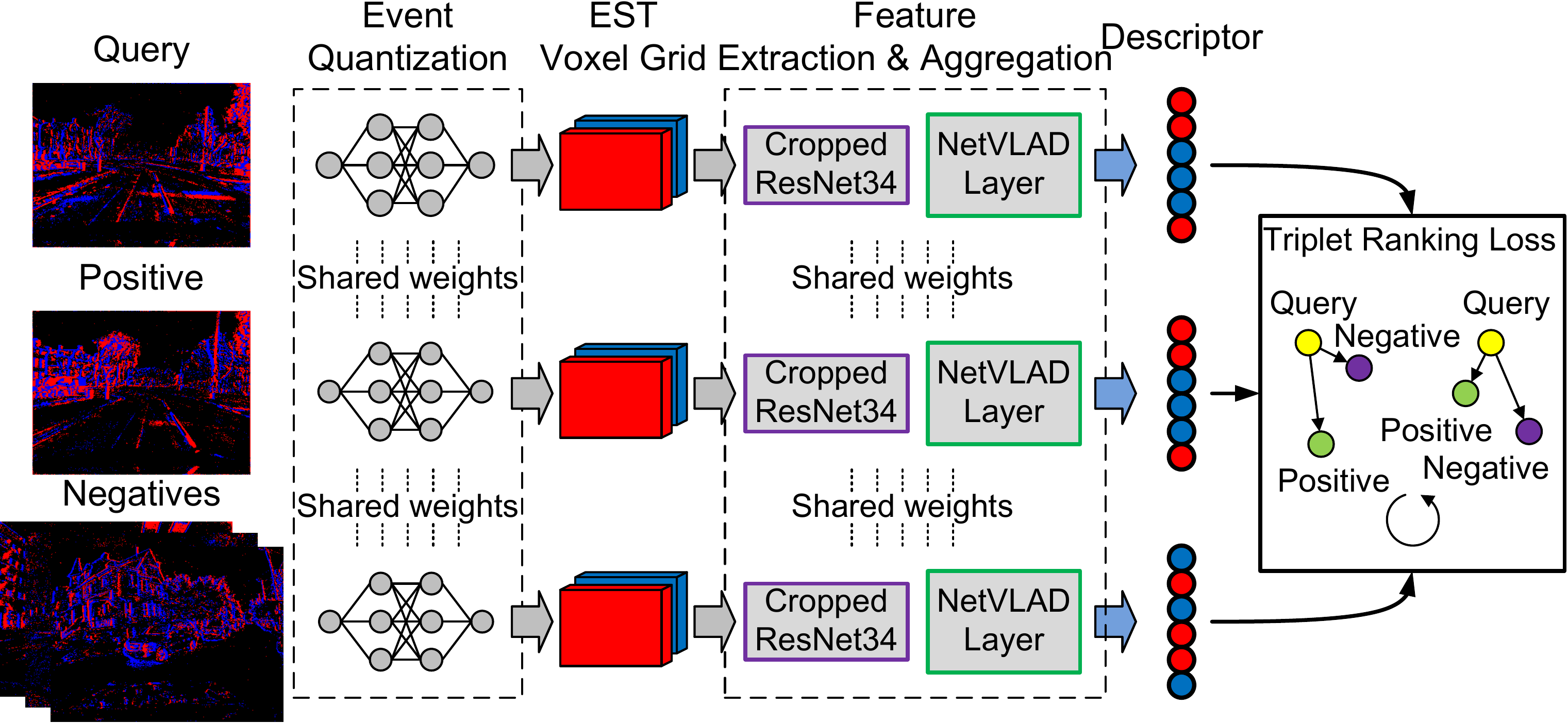}
\caption{Overview of Our Event-VPR Architecture. We propose a novel end-to-end event-based VPR network. For a given event bin of query, our method selects the corresponding positive and negatives of event bins, and trains the network model through the triplet ranking loss to learn the global descriptor vectors of the event bins.}
\label{fig:1}
\vspace{-3mm}
\end{figure} 

In contrast to traditional frame-based VPR methods, we propose a novel VPR method using event cameras in this paper. Event cameras are neuromorphic visual sensors inspired by the biological retina, and work in a completely different way from frame-based cameras: they use address-event representation (AER) and output pixel-level brightness changes (called events) with microsecond resolution to generate sparse asynchronous event streams. Event cameras have the advantages of low latency, high temporal resolution, low bandwidth, low power consumption and high dynamic range \cite{gallego2020event}, which can effectively overcome the problems existing in typical frame-based cameras. To achieve robust VPR using event cameras, we propose a novel \textbf{end-to-end visual place recognition network for event cameras (Event-VPR)}. The key idea, as shown in Fig. \ref{fig:1}, is to apply NetVLAD to EST voxel grid representation generated by event streams. To the best of our knowledge, this is the first end-to-end VPR method for event cameras. Experimental results on multiple datasets with different weather and scenes show that the proposed method is superior to the traditional frame-based VPR methods, and can effectively solve the challenges of large-scale scenes, high dynamic range and long-term adaptability in visual place recognition. 

The main contributions of this paper are as follows:
\begin{itemize}
\item A novel end-to-end weakly supervised network pipeline of visual place recognition for event cameras is proposed, which directly uses event streams as input.
\item Different event representations, network structures and parameters of the proposed network are compared to show how it affects the total performance.
\item The performance of the proposed method against frame-based methods on different datasets with different weather and seasons is thoroughly evaluated and the source code of our method is open-sourced\footnote{\url{https://github.com/kongdelei/Event-VPR}}. 
\end{itemize}


\section{RELATED WORK}
\label{sec:relatedwork}
Visual sensors are the main sensor types for place recognition due to their low cost, low power consumption and abundant scene information. Nowadays, most popular VPR methods generally use frame-based visual sensors and appearance-based \cite{oertel2020augmenting} approaches to realize large-scale place recognition. In this case, the VPR problem can generally be transformed into a large-scale geo-tagged image retrieval problem, and that can be solved by image matching. Extensive researches on how to represent and match images have been carried out \cite{lowry2015visual} \cite{zeng2018place}. Those methods usually use traditional feature extraction techniques (such as SIFT \cite{lowe2004distinctive} and ORB \cite{rublee2011orb}), and local descriptor aggregation techniques (such as BoW \cite{angeli2008fast} \cite{galvez2012bags} and VLAD \cite{jegou2010aggregating} \cite{arandjelovic2013all}) to establish a higher-order statistical model of image features. With the rise of deep learning, some works begin to use off-the-shelf convolutional networks (such as Overfeat \cite{chen2014convolutional}, VggNet and AlexNet \cite{lopez2017appearance}) as trainable feature extractors. Some researchers also revised VLAD into a trainable pooling layer (such as NetVLAD \cite{arandjelovic2016netvlad} and DenseVLAD \cite{torii201524}) to obtain image descriptor vectors as compact image representations. In the retrieval and matching process, sequence-based matching is a widely recognized method. The most known work is SeqSLAM \cite{milford2012seqslam}, which searches highly similar short image sequences for VPR. Recently, researchers also tried to further improve the recognition performance from different aspects. For example, most structure-based methods use structural information such as repeated edges and semi-dense maps of the scene \cite{oertel2020augmenting,torii2013visual,ye2017place} for place recognition. In recent years, there are also some semantic-based works for VPR that mainly use semantic information such as landmarks, texts and objects in the scene. \cite{camara2019spatio,hong2019textplace,benbihi2020image}.

Though traditional frame-based VPR methods have developed rapidly over the past decade, they are still difficult to solve some problems in challenging scenarios (such as illumination changes and motion blur) due to the inherent defects of frame-based cameras. Compared to standard frame-based cameras, event cameras have many advantages such as high dynamic range, high temporal resolution and low latency \cite{gallego2020event}. Due to the above advantages, event cameras have drawn more and more attention. However, to the best of our knowledge, there are still few works related to the event-based VPR problem. Among them, Milford et al. first tried to migrate SeqSLAM to the event camera in 2018 and completed a relatively basic recognition experiment based on event frames \cite{milford2015place}. Recently, they proposed an event-based VPR scheme with ensembles of spatio-temporal windows (Ensemble-Event-VPR) \cite{fischer2020event}. This method uses event bins with different numbers of events and sizes of temporal windows to reconstruct a group of intensity frame sequences using E2Vid \cite{rebecq2019events} for place recognition. However, this method is not a direct event-based method but needs to convert the events into intensity frames, which is still a frame-based VPR method in essence. Different from that, in this paper, we propose a novel end-to-end event-based visual place recognition network (Event-VPR) that directly uses event streams, which achieves excellent recognition results even in challenging environments.

\section{METHODOLOGY}
\label{sec:methodology}
In this section, we will describe the network architecture of Event-VPR in detail, including the various module components of our algorithm and the main steps of network training.

\begin{figure*}[t]
\centering
\includegraphics[width=1.95\columnwidth]{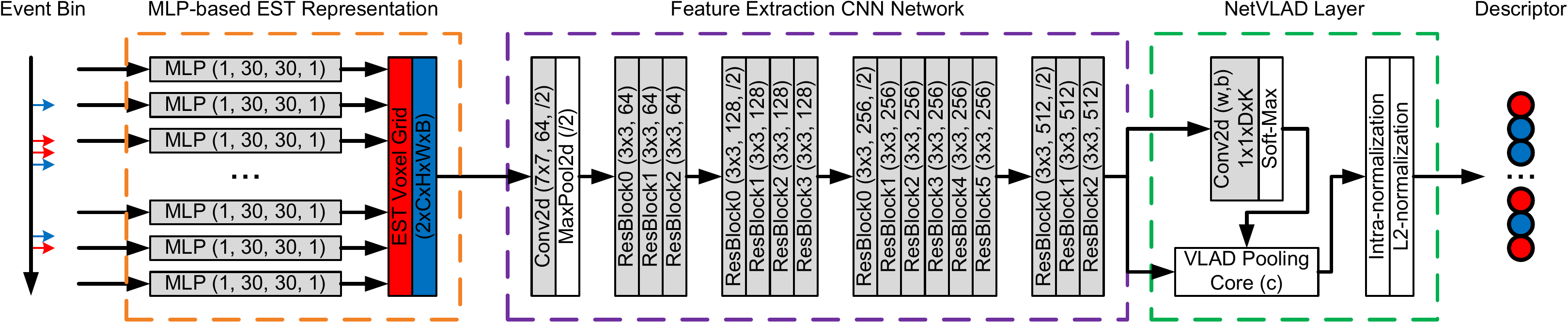}
\caption{Overview of Our Proposed Pipeline. Firstly, the event bins are converted into EST voxel grids by MLP-based kernel. Then, the cropped ResNet34 convolution network is used to extract visual features of EST voxel grids. Next, VLAD-based local aggregated description layer is used for feature descriptor aggregation. Finally, the triplet ranking loss is used to train the network with weakly supervised training.}
\label{fig:2}
\vspace{-3mm}
\end{figure*}

\subsection{Problem and Pipeline}
\textbf{Problem Definition.}
Define a database of events $D=\{P, E\}$ that contains $n$ geo-location coordinated $\boldsymbol{P}=\left\{P_{1}, \cdots, P_{n}\right\}$ under a fixed reference system and $n$ groups of event bins $\boldsymbol{E}=\left\{E_{1}, \cdots, E_{n}\right\}$, where each place coordinate $P_{i}$ corresponds to several event bins $E_{i}=\left\{E_{i 1}, \cdots, E_{i m}\right\}$ and each event bin $E_{i j}$ is collected by using an event camera in the area near the place coordinate $P_{i}$. The area of coverage (AOC) of all sub-areas is approximately the same, i.e. $\operatorname{AOC}\left(P_{1}\right) \approx \cdots \approx \operatorname{AOC}\left(P_{n}\right)$.
The problem of event-based VPR can be formally defined as follows: Given a query event bin denoted as $E_{q}$, our goal is to retrieve several event bins from the database $D$ that are most similar to $E_{q}$, thus obtaining their corresponding geographic location coordinates $P_{q}$. To this end, we design a deep neural network to learn a function $f_{\text {Event-VPR }}(\cdot)$, which is used to map a given event bin $E_{q}$ to a global descriptor vector $v_{q}=f\left(E_{q}\right)$ such that $\mathrm{d}\left(v_{q}, v_{r}\right)<\mathrm{d}\left(v_{q}, v_{s}\right)$ if $v_{q}$ is similar to $v_{r}$ but different from $v_{s}$. Here $\mathrm{d}(\cdot)$ is a distance function (such as euclidean distance function). Our problem then is simplified to find the place coordinates $P_{q}$ of the sub-area such that the global descriptor vector $v_{*}$ from one of its event bin gives the minimum distance with the global descriptor vector $v_{q}$ from the query, i.e. $\mathrm{d}\left(v_{q}, v_{*}\right)<\mathrm{d}\left(v_{q}, v_{i}\right), \forall i \neq *$.

\textbf{Our Pipeline.}
The event-based visual place recognition (Event-VPR) proposed in this paper is a novel end-to-end VPR method using event-based camera. The key idea of the algorithm is as follows: Firstly, we divide consecutive event stream into event bins and convert event bins into EST voxel grid representations using MLP-based kernel. Then, a cropped ResNet34 network \cite{he2016deep} is used to extract visual features of EST voxel grids. Next, a VLAD-based local aggregated description layer is used for feature descriptor aggregation. Finally, the triplet ranking loss is used to train the network with weakly supervised training. Corresponding to the aforementioned key idea, our pipeline is mainly divided into the following 4 parts: EST voxel grid representation, feature extraction convolution network, feature aggregated description layer and triplet ranking loss. Its architecture is shown in Fig. \ref{fig:2}.

\subsection{Events and EST Voxel Grid} 
\textbf{Event-based Data.}
The pixel array of the event camera is capable of independently and logarithmically responding to pixel-level brightness changes (i.e. $L \doteq \log (I)$, where $I$ is photocurrent) and triggering sparse asynchronous events $E=\left\{e_{1}, \cdots, e_{\mathrm{N}} \mid e_{k} \in \mathbb{R}^{4}\right\}$. To be exact, without considering fixed pattern noise (FPN), the brightness change at the pixel $\left(x_{k}, y_{k}\right)^{\mathrm{T}}$ at time $t_{k}$ is given by:
\begin{equation}
\label{eq:1}
\begin{array}{l}
\Delta L\left(x_{k}, y_{k}, t_{k}\right)=L\left(x_{k}, y_{k}, t_{k}\right)-L\left(x_{k}, y_{k}, t_{k}-\Delta t_{k}\right) \\
\left|\Delta L\left(x_{k}, y_{k}, t_{k}\right)\right| \geq \vartheta
\end{array}
\end{equation}
where $\Delta t_{k}$ is the time interval between last triggered event and current triggered event of a pixel. When the brightness change of a pixel reaches the contrast threshold $\vartheta$ (here $\vartheta>0$), the pixel triggers an event $e_{k}=\left(x_{k}, y_{k}, t_{k}, p_{k}\right)^{\mathrm{T}}$. Here, $p_{k}$ is event polarity which is given by:
\begin{equation}
\label{eq:2}
p_{k}=\frac{\Delta L\left(x_{k}, y_{k}, t_{k}\right)}{\left|\Delta L\left(x_{k}, y_{k}, t_{k}\right)\right|}=\{-1,1\}
\end{equation}
In real sensor, positive events (ON) and negative events (OFF) can be triggered according to different contrast thresholds $\vartheta$. Furthermore, the events can be summarized as an event measurement field with polarity defined on the 3D continuous spatio-temporal manifold:
\begin{equation}
\label{eq:3}
S(x, y, t)=\sum_{e_{k} \in E} p_{k} \delta\left(x-x_{k}, y-y_{k}\right) \delta\left(t-t_{k}\right)
\end{equation}
where $\delta(\cdot)$ denotes the dirac pulse defined in the event domain and is used to replace each event.

\begin{figure}[htbp]
\centering
\includegraphics[width=0.95\columnwidth]{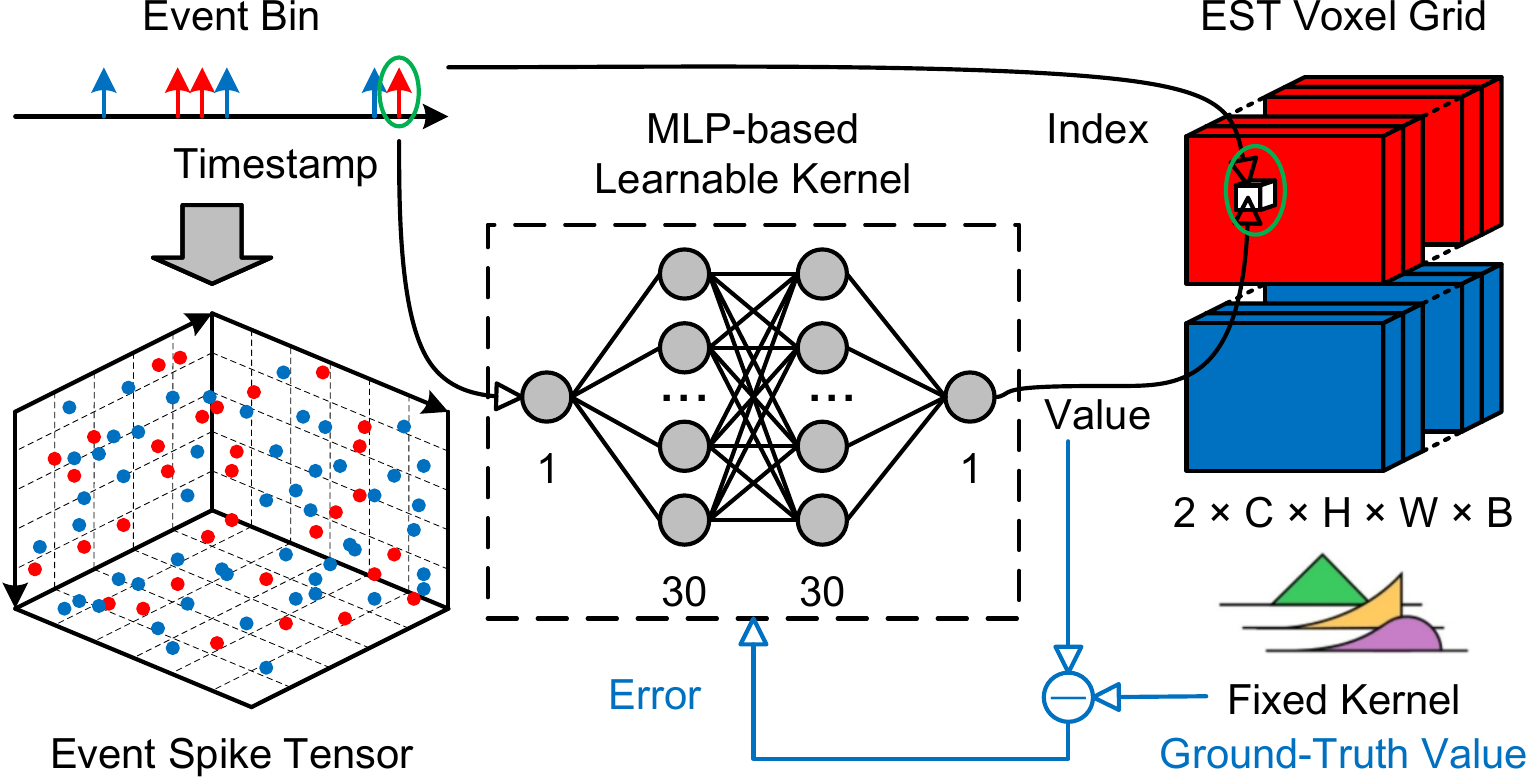}
\caption{Overview of EST Voxel Grid for Event-based Data. The value generated by MLP-based kernel for each event is put into the corresponding coordinates, and kernel is learnable through end-to-end training.}
\label{fig:3}
\vspace{-3mm}
\end{figure}

\textbf{EST Voxel Grid.}
The output of an event camera is the consecutive event stream. In order to use deep neural networks to extract features from event stream, we need to convert the event streams into a representation that convolutional network could process. Nowadays, typical representation methods of events roughly include motion-compensated event frame (EF) \cite{gallego2018unifying}, 4-channel event count and last-timestamp image (4CH) \cite{zhu2018ev}, and event spike tensor (EST) \cite{gehrig2019end}, etc. In addition, the events can also be converted into traditional frame-based video (e.g. E2Vid) \cite{rebecq2019events}. In this paper, we firstly divide event stream into event bins, then we use the voxel grid of event spike tensor (EST) to represent event bins. In order to obtain most meaningful visual feature information from the event measurement field, we convolve it with a trilinear voting kernel $k(x, y, t)$. Therefore, the convolution signal becomes:
\begin{equation}
\label{eq:4}
\begin{array}{l}
(k * S)(x, y, t)=\sum\limits_{e_{k} \in E} p_{k} k\left(x-x_{k}, y-y_{k}, t-t_{k}\right) \\
k(x, y, t)=\delta(x, y) \max \left(0,1-\left|\frac{t}{\Delta t}\right|\right)
\end{array}
\end{equation}
After the kernel convolution, the signal (\ref{eq:4}) can be periodically sampled on the spatio-temporal coordinates:
\begin{equation}
\label{eq:5}
\begin{array}{l}
V_{\mathrm{EST}}\left[x_{l}, y_{m}, t_{n}\right] \\
=\sum\limits_{e_{k} \in E} p_{k} \delta\left(x_{l}-x_{k}, y_{m}-y_{k}\right) \max \left(0,1-\left|\frac{t_{n}-t_{k}}{\Delta t}\right|\right)
\end{array}
\end{equation}
where $\left(x_{l}, y_{m}, t_{n}\right)$ are the sampled grid space-time coordinates, with $x_{l} \in\{0,1, \ldots, W-1\}$, $y_{m} \in\{0,1, \ldots, \mathrm{H}-1\}$, $t_{n} \in\left\{t_{0}, t_{0}+\Delta t, \ldots, t_{0}+(\mathrm{C}-1) \Delta t\right\}$. Here, $(\mathrm{W}, \mathrm{H})$ is the pixel size, $t_{0}$ is the start timestamp, $\Delta t$ is the size of time blocks, $\mathrm{C}$ is the number of time blocks (that is, the number of channels).
We replace the manually designed kernel in (\ref{eq:5}) with a multi-layer perceptron (MLP) to generate EST voxel grid as an end-to-end events representation, as shown in Fig. \ref{fig:3}, where $\mathrm{C}$ is the number of channels, $\mathrm{B}$ is the batch size. The MLP receives the normalized timestamp of the event as input and has 2 hidden layers, each with 30 neurons. The value generated by MLP for each event is put into the corresponding voxel grid coordinates. EST voxel grid can be optimized according to specific tasks to optimize the performance of the whole network.

\subsection{Feature Extraction and Aggregation}
\textbf{Feature Extraction Network.}
To extract features from the EST voxel grid, we use the ResNet34 pre-trained in event-based hand-written number recognition tasks \cite{he2016deep}. In order to migrate it to our VPR task, the original network needs to be cropped. We modify the number of input channels of the first convolution layer of ResNet34 to make it suitable for EST voxel grid as network input. Meanwhile, in order to connect the feature description aggregation layers, we remove the last average pooling layer and the full connection layer at the end of the ResNet34. The network output $\mathrm{x}=f_{\text{ResNet34}}\left(V_{\mathrm{EST}}\right)$ is the feature maps of the EST voxel grid. Here, $\mathrm{x}$ is the $\mathrm{w} \times \mathrm{h} \times \mathrm{D}$-dimensional feature tensor obtained by the feature extraction network. For ResNet34, $\mathrm{w} \times \mathrm{h} = 7 \times 7$.

\begin{figure}[htbp]
\centering
\includegraphics[width=0.95\columnwidth]{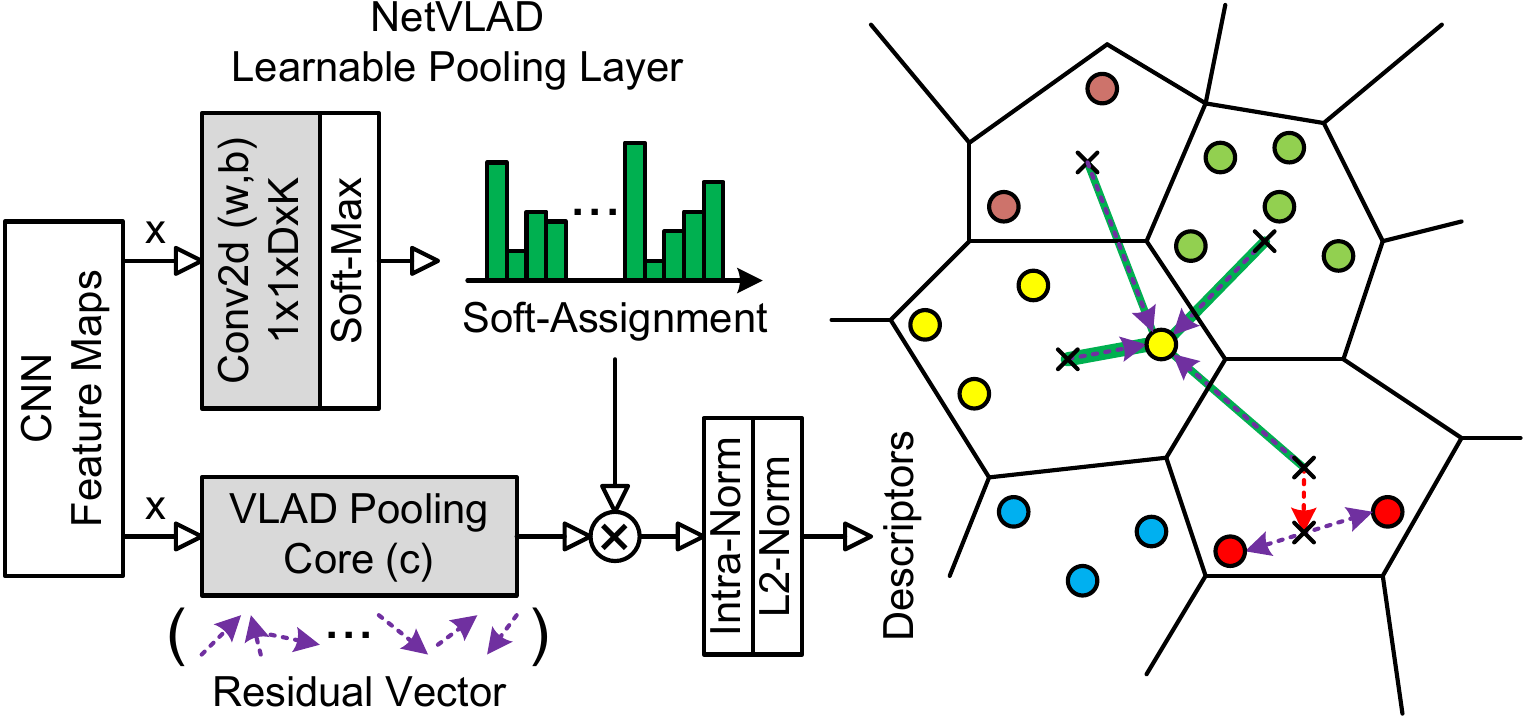}
\caption{Overview of VLAD-based Pooling Layer. Residual vector and soft assignment are statistics from descriptors to clusters, and the convolution layer and clusters are learnable through end-to-end training.}
\label{fig:4}
\vspace{-3mm}
\end{figure}

\textbf{Feature Aggregation Network.}
After getting features for the EST voxel grid, we need to aggregate the features for descriptor matching. We use vector of locally aggregated descriptor (VLAD) \cite{jegou2010aggregating} \cite{arandjelovic2013all} which is a trainable descriptor pooling method commonly used in the fields of image retrieval. As shown in Fig. \ref{fig:4}, we interpret the $\mathrm{w} \times \mathrm{h} \times \mathrm{D}$-dimensional feature maps $\mathrm{x}$ output by the feature extraction network as $\mathrm{M}$ $\mathrm{D}$-dimensional local descriptors $\left\{\mathrm{x}_{1}, \cdots, \mathrm{x}_{\mathrm{M}} \mid \mathrm{x}_{i} \in \mathbb{R}^{\mathrm{D}}\right\}$ as input, and $\mathrm{K}$ $\mathrm{D}$-dimensional cluster centers $\left\{\mathrm{c}_{1}, \cdots, \mathrm{c}_{\mathrm{K}} \mid \mathrm{c}_{k} \in \mathbb{R}^{\mathrm{D}}\right\}$ as VLAD parameters. The normalized output of descriptor vector $V_{\mathrm{VLAD}}$ is a $\mathrm{D} \times \mathrm{K}$-dimensional matrix, which is given by:
\begin{equation}
\label{eq:6}
V_{\mathrm{VLAD}, k}(\mathrm{x})=\sum_{i=1}^{\mathrm{M}} \bar{a}_{k}\left(\mathrm{x}_{i}\right)\left(\mathrm{x}_{i}-c_{k}\right)
\end{equation}
where $\left(\mathrm{x}_{i}-c_{k}\right)$ is the residual vector of descriptor $\mathrm{x}_{i}$ to cluster center $\mathrm{c}_{k}$, and $\bar{a}_{k}\left(\mathrm{x}_{i}\right)$ denotes the soft assignment of descriptor $\mathrm{x}_{i}$ to cluster center $\mathrm{c}_{k}$, which is given by:
\begin{equation}
\label{eq:7}
\bar{a}_{k}\left(\mathrm{x}_{i}\right)=\frac{e^{-\alpha\left\|\mathrm{x}_{i}-\mathrm{c}_{k}\right\|^{2}}}{\sum_{k^{\prime}} e^{-\alpha\left\|\mathrm{x}_{i}-\mathrm{c}_{k^{\prime}}\right\|^{2}}}=\frac{e^{\mathrm{w}_{k}^{\mathrm{T}} \mathrm{x}_{i}+b_{k}}}{\sum_{k^{\prime}} e^{\mathrm{w}_{k^{\prime}}^{\mathrm{T}} \mathrm{x}_{i}+b_{k^{\prime}}}}
\end{equation}
It assigns the weight of descriptor $\mathrm{x}_{i}$ to cluster center $\mathrm{c}_{k}$ according to their proximity distance. Where, $\mathrm{w}_{k}=2 \alpha \mathrm{c}_{k}$, $b_{k}=-\alpha\left\|\mathrm{c}_{k}\right\|^{2}$, $\alpha>0$, and $\bar{a}_{k}\left(\mathrm{x}_{i}\right) \in[0,1]$.
According to (\ref{eq:7}), soft assignment can be decomposed into a convolution layer and a soft-max layer, and the weight $\left\{\mathrm{w}_{k}\right\}$ and bias $\left\{b_{k}\right\}$ of the convolution layer are taken as independent trainable parameters together with the clustering center $\left\{\mathrm{c}_{k}\right\}$.
Finally, the aggregated vector needs to be intra-normalized and L2-normalized to produce the final global descriptor vector $v \in \mathbb{R}^{\Omega}$, $\|v\|_{2}=1$, $\Omega=\mathrm{D} \times \mathrm{K}$ for event bins that can be used for efficient retrieval.

\subsection{Network Training}
\textbf{Training Triplet Building.}
We use the metric learning to train Event-VPR in an end-to-end manner to learn the function $f_{\text {Event-VPR}}(\cdot)$ that maps the input of event bins $E$ into global descriptor vectors $v \in \mathbb{R}^{\Omega}$. In the process of network training, the training query $E_{q}$ and its geographical location $P_{q}$ is given. It is necessary to select suitable best positive and hard negatives from the database $\boldsymbol{D}=\{\boldsymbol{P}, \boldsymbol{E}\}$. In order to improve the search efficiency, firstly, all samples within the range of geographical distance $\lambda$ are selected as potential positives $\left\{E_{p o s, i}\right\}$ according to the location $P_{q}$ of the query:
\begin{equation}
\label{eq:8}
\mathrm{d}_{geo}\left(P_{q}, P_{pos, i}\right) \leq \lambda, \forall E_{pos, i} \subseteq \boldsymbol{E}
\end{equation}
Then, among these potential positives, the positive with the smallest descriptor vector distance is selected as the best positive, and its descriptor vector is:
\begin{equation}
\label{eq:9}
v_{best\text{-}pos}=\underset{i}{\arg \min } \mathrm{d}_{\gamma}\left(v_{q}, v_{pos, i}\right)
\end{equation}
When selecting negatives, firstly, all samples outside the range of geographical distance $\delta$ are selected according to the location $P_{q}$ of the query, and then $n_{sample}$ of them are selected as randomly sampled negatives $\left\{E_{neg, i}\right\}$:
\begin{equation}
\label{eq:10}
\mathrm{d}_{geo}\left(P_{q}, P_{neg, i}\right) \geq \delta, \forall E_{neg, i}, \subseteq \boldsymbol{E}
\end{equation}
Then, among these randomly sampled negatives, the samples that violates the margin condition are selected as the candidate hard negatives $\left\{E_{hard\text{-}neg, i}\right\}$:
\begin{equation}
\label{eq:11}
\mathrm{d}_{\gamma}^{2}\left(v_{q}, v_{hard\text{-}neg, i}\right) \leq \mathrm{d}_{\gamma}^{2}\left(v_{q}, v_{pos, j}\right)+m, \forall E_{hard\text{-}neg, i} \subseteq \boldsymbol{E}
\end{equation}
where $m$ is a constant parameter, representing the margin between $\mathrm{d}_{\gamma}\left(v_{q}, v_{pos, i}\right)$ and $\mathrm{d}_{\gamma}\left(v_{q}, v_{neg, j}\right)$. In order to improve the training efficiency, we select the $n_{neg}$ samples with the smallest descriptor vector distance as the hard negatives for training, where $n_{neg} \in\left[0, \mathrm{N}_{neg}\right]$, $n_{neg} \ll n_{sanple}$.

\textbf{Loss Function.}
Based on the above method, we use the data from the event camera datasets to obtain a set of training tuples from the training set, where each triple is represented as $\xi=\left(E_{q}, E_{best\text{-}pos},\left\{E_{hard\text{-}neg, j}\right\}\right)$.
Here, $E_{q}$ is the query, $E_{best\text{-}pos}$ is the best positive, and $\left\{E_{hard\text{-}neg,j}\right\}$ is a group of hard negatives. If their global descriptor vector is $\gamma=\left(v_{q}, v_{best\text{-}pos},\left\{v_{hard\text{-}neg, j}\right\}\right)$, the descriptor vector distances of query $E_{q}$ to the best positive $E_{best\text{-}pos}$ and the hard negatives $\left\{E_{hard\text{-}neg,j}\right\}$ are defined as $\mathrm{d}_{\gamma}\left(v_{q}, v_{best\text{-}pos}\right)$ and $\mathrm{d}_{\gamma}\left(v_{q}, v_{hard\text{-}neg, j}\right)$ respectively. The loss function is designed to minimize the global descriptor distance between the query and the best positive, and to maximize the distance between the query and the hard negatives. We use weakly supervised triplet ranking loss which is defined as follows:
\begin{equation}
\label{eq:12}
\begin{array}{l}
L_{\text {triplet }}(\gamma)    \\
=\sum_{j}\left[\mathrm{d}_{\gamma}\left(v_{q}, v_{best\text{-}pos}\right)-\mathrm{d}_{\gamma}\left(v_{q}, v_{hard\text{-}neg, j}\right)+m\right]_{+}
\end{array}
\end{equation}
where $m$ denotes the margin, and $[\cdot]_{+}=\max (\cdot, 0)$ means the loss takes a positive number, i.e. $L_{\text {triplet }}(\gamma) \geq 0$.

\section{EXPERIMENTS}
\label{sec:experiments}
In this section, we carried out experiments on multiple datasets, including the MVSEC\cite{zhu2018multivehicle}, DDD17\cite{binas2017ddd17} and Oxford RobotCar datasets \cite{maddern20171} to verify the effectiveness of the proposed method through quantitative and qualitative experimental results (A detailed description of the experimental configuration is given in APPENDIX). We conducted four experiments to evaluate our proposed method. Firstly, we evaluated the performance of the Event-VPR in different driving scenes, and verified its long-term robustness. Then, we compared the Event-VPR with frame-based VPR algorithms on the same dataset, and provided numerical analysis to show the performance of Event-VPR. Finally, we performed network ablation study on Event-VPR from two aspects (event representation and feature extraction network), and the experimental results proved the advantages of each module in our method.

\subsection{Performance Comparison in Different Scenarios}
\begin{figure}[h]
\centering
\includegraphics[width=0.95\columnwidth]{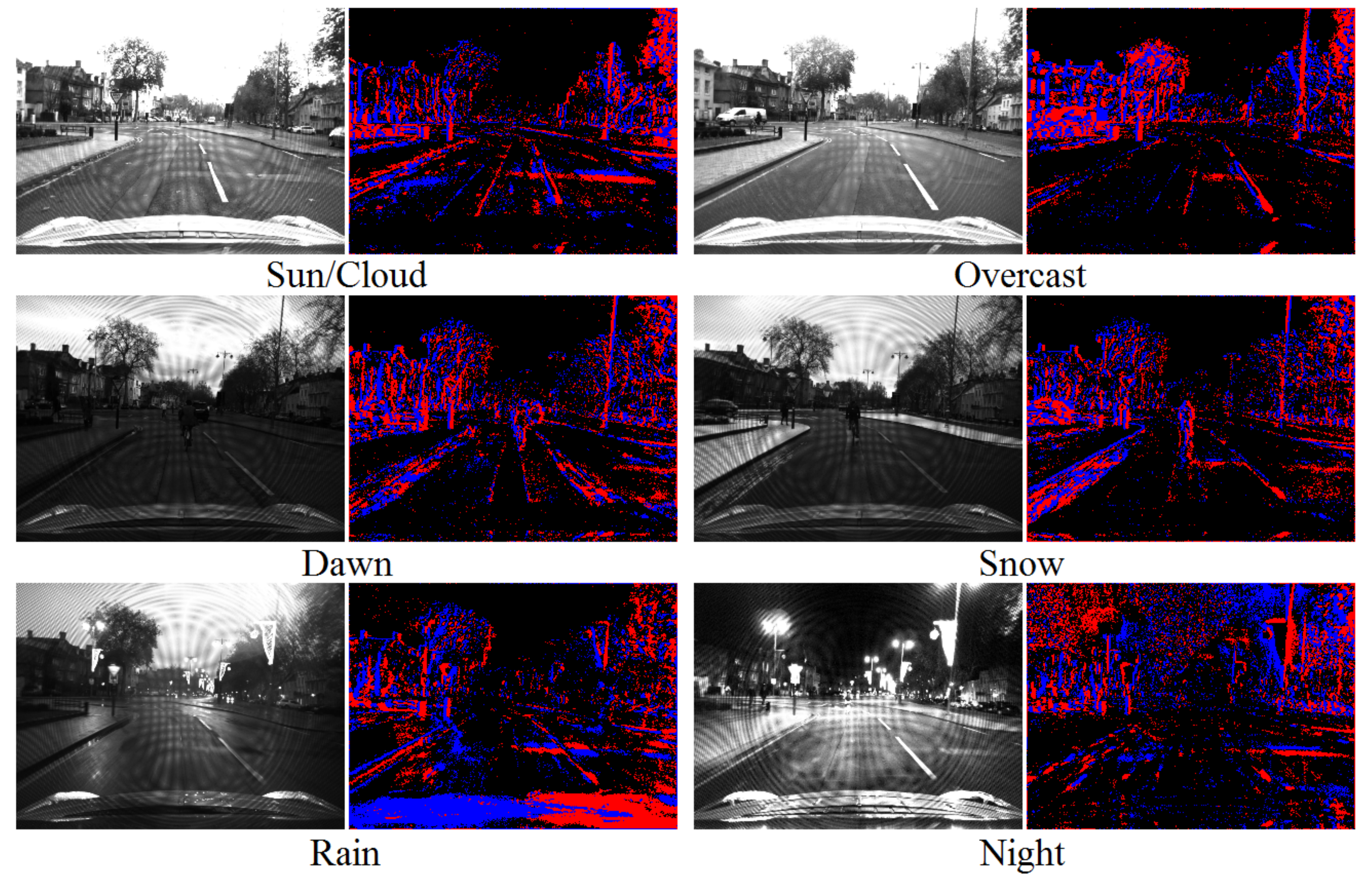}
\caption{Synthetic Event Bins of Oxford RobotCar Datasets using V2E \cite{delbruck2020v2e}.}
\label{fig:5}
\vspace{-3mm}
\end{figure}

In the first experiment, we validated the performance of our Event-VPR in different scenarios of MVSEC, DDD17 and Oxford RobotCar datasets. 
We selected several challenging sequences for our experiments. For example, the night scene sequence in the MVSEC dataset is dim with little light. There are many similar scenes and high dynamic range scenes (e.g. clouds and tunnels) in the expressway scenes in the DDD17 dataset. As shown in Fig. \ref{fig:5}, there are also changing weather and seasonal conditions in the Oxford RobotCar dataset. For instance, there are light reflection from the surface of water on rainy day and the bright street lamps at night, which result in a high dynamic range scenario for visual cameras. 

\textbf{Results on MVSEC.}
As shown in Table \ref{tab:1}, the experimental results on the MVSEC dataset show the recognition performance (Recall@N) of the method in daytime and nighttime scenes. On the MVSEC dataset, we trained 2 daytime and 3 nighttime sequences together, and then tested each sequence separately. The results show that the Recall@1 of our model in night sequences can achieve 97.05\% on average, which is almost the same as that in the daytime sequences (Please refer to APPENDIX to see how we select training and test sets).
\begin{table}[htbp]
\begin{center}
\caption{Recalls of Event-VPR(EST) on MVSEC Dataset}
\setlength{\tabcolsep}{1pt}
\newcommand{\tabincell}[2]{\begin{tabular}{@{}#1@{}}#2\end{tabular}}
\begin{tabular*}{0.95\linewidth}{@{}@{\extracolsep{\fill}}c|ccccp{1cm}@{}}
\toprule
Suquences   & Recall@1  & Recall@5  & Recall@10 & Recall@20    \\
\midrule
All-day1    & 99.51\%         & 99.84\%         & 100.00\%        & 100.00\%    \\
All-day2    & 91.52\%         & 97.57\%         & 99.17\%         & 99.58\%     \\
All-night1  & 98.67\%         & 99.33\%         & 100.00\%        & 100.00\%    \\
All-night2  & 95.11\%         & 98.22\%         & 99.11\%         & 100.00\%     \\
All-night3  & 97.37\%         & 98.68\%         & 99.34\%         & 100.00\%     \\
\bottomrule
\end{tabular*}
\label{tab:1}
\vspace{-8mm}
\end{center}
\end{table}

\begin{figure*}[htbp]
\centering
	\subfigure{
	    \label{all}
		\includegraphics[width=0.64\columnwidth]{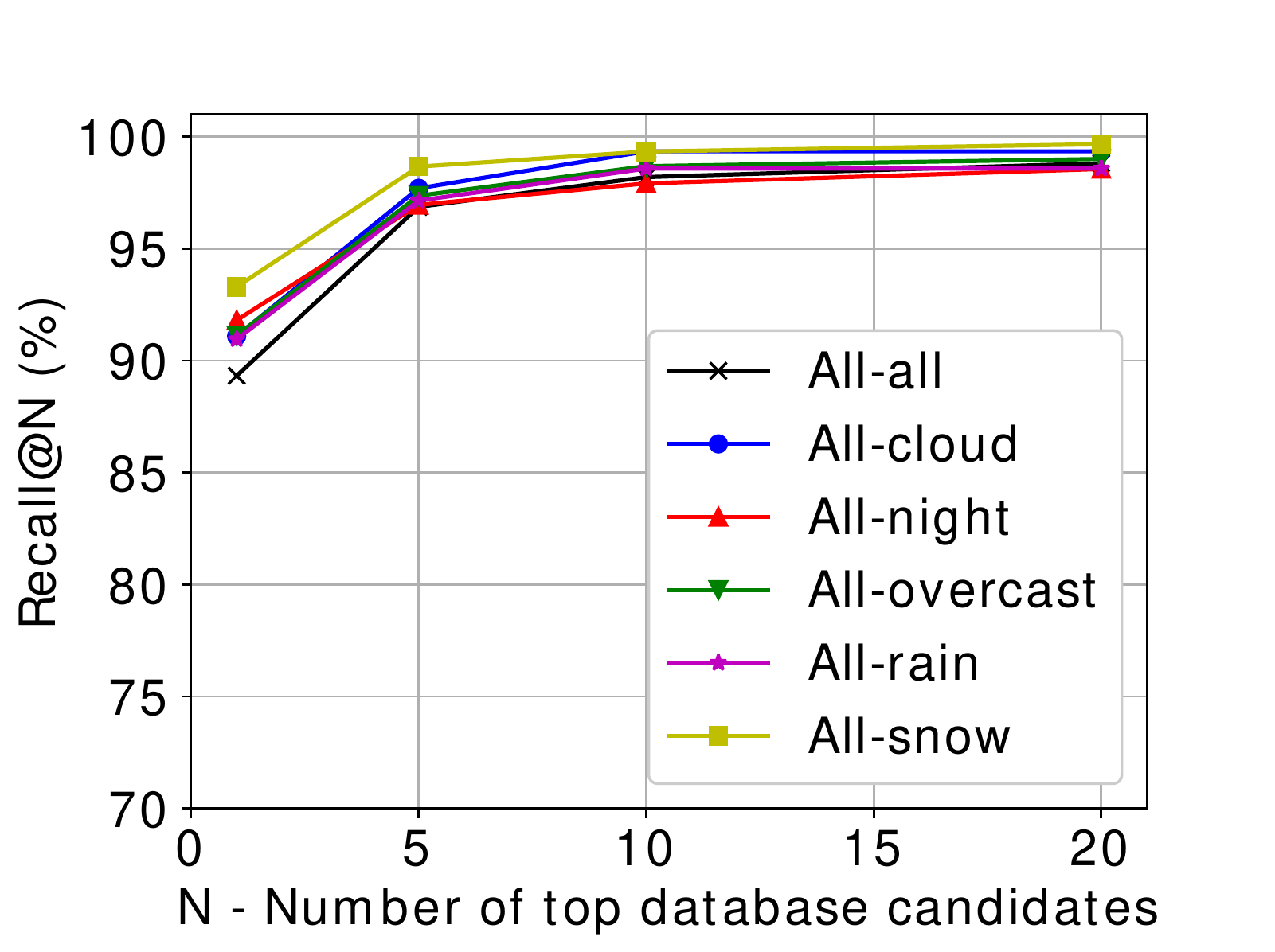}
	}
	\hspace{0.000001\columnwidth}
	\subfigure{
    	\label{day}
		\includegraphics[width=0.64\columnwidth]{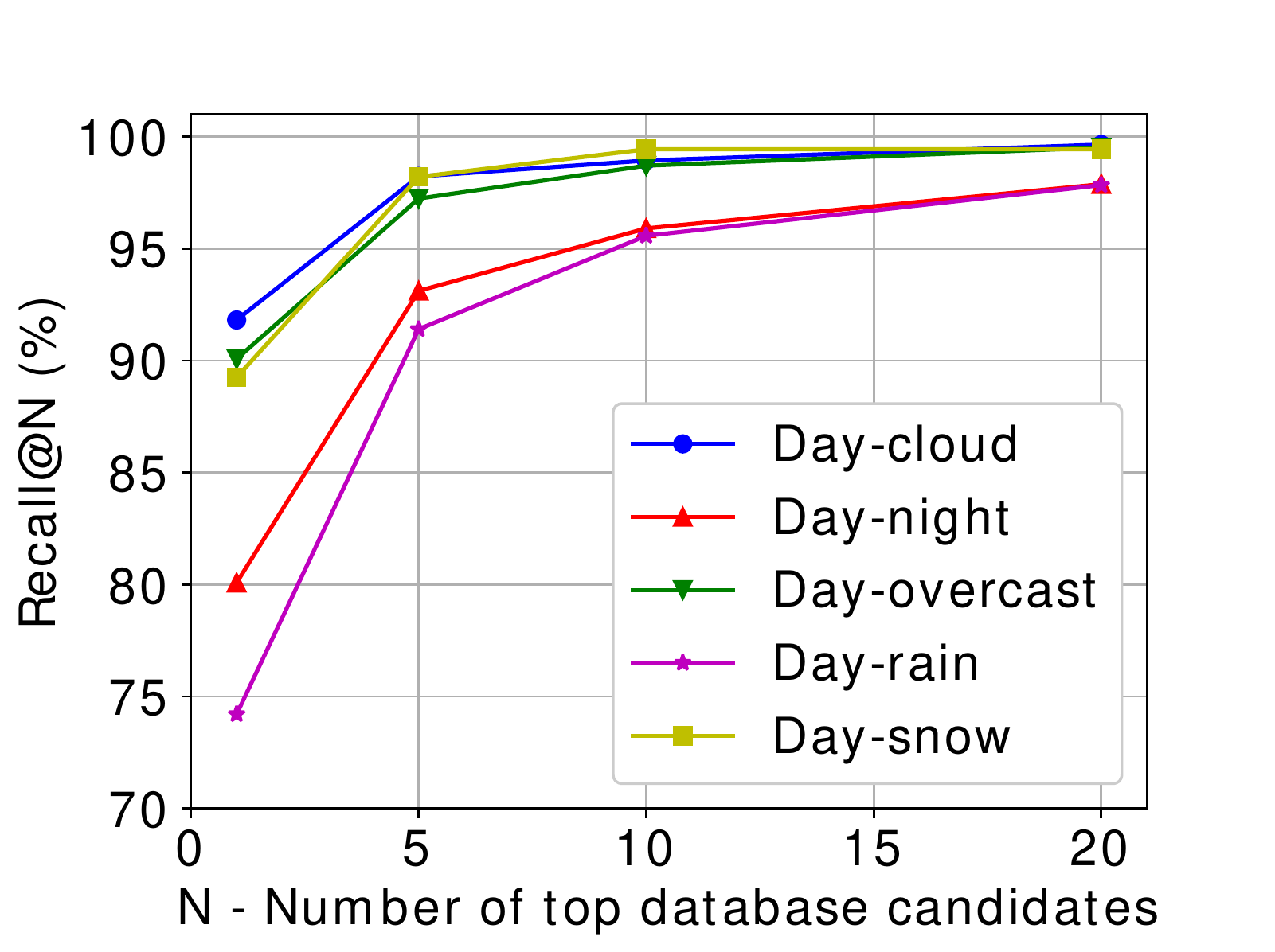}
	}
	\hspace{0.000001\columnwidth}
	\subfigure{
	    \label{night}
		\includegraphics[width=0.64\columnwidth]{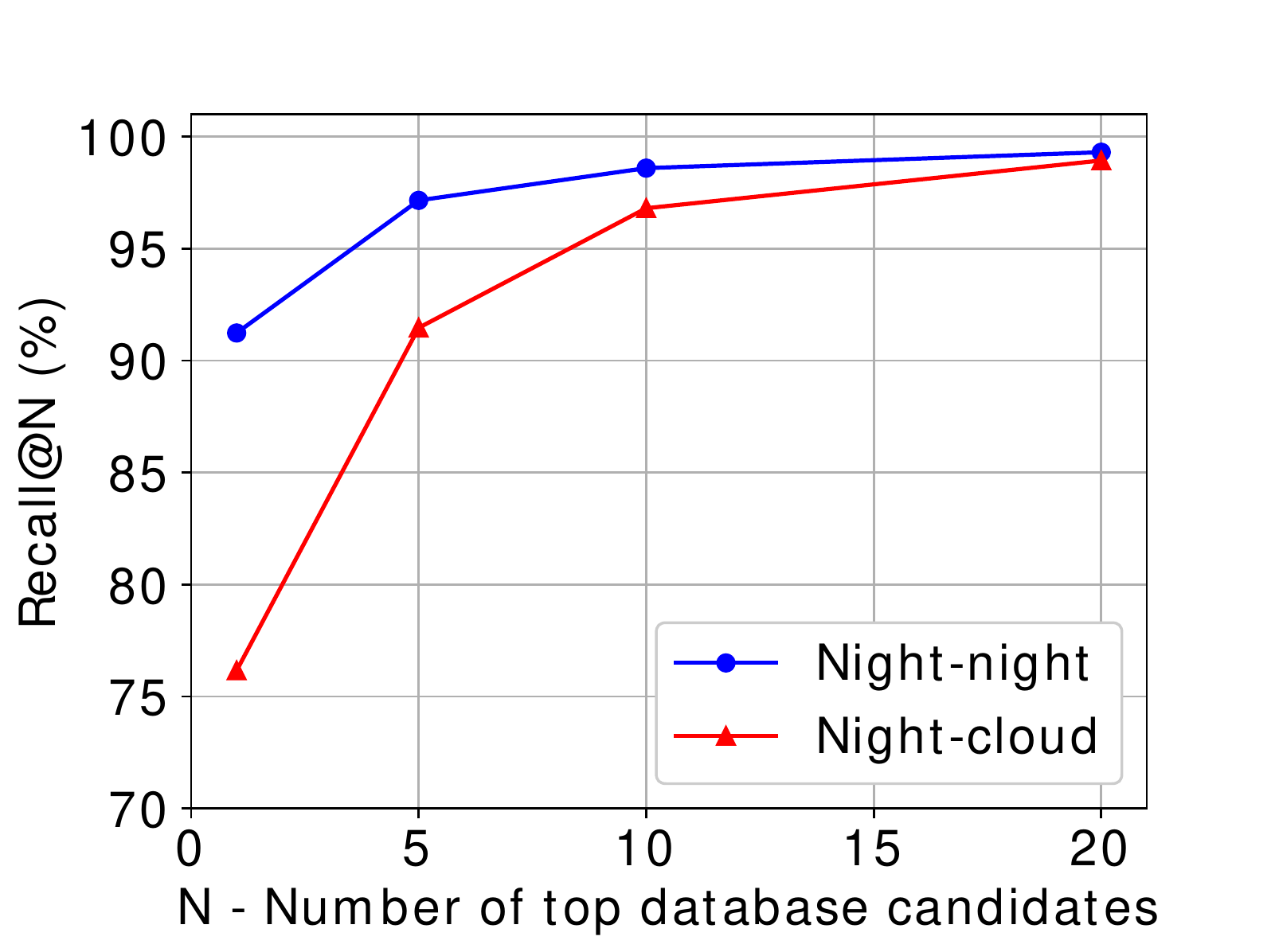}
	}
\caption{Recalls of Event-VPR on Day-Night Sequences (Oxford RobotCar Dataset). From left to right: (a) All, (b) Day, (c) Night.}
\label{fig:6}
\vspace{-5mm}
\end{figure*}

\textbf{Results on Oxford RobotCar.}
Fig. \ref{fig:6} shows the recognition performance of our method under various weather and seasons on the Oxford RobotCar dataset. The experimental results of model trained on all sequences and tested on different weather sequences are shown in Fig. \ref{all} (Please refer to APPENDIX for details). Our model can achieve high performance in different weather conditions. We trained models in day and night sequences separately, and tested on untrained scene sequences (e.g. overcast, rain and snow). The experimental results of model trained on all daytime sequences (including sun and cloud) but tested on different weather sequences are shown in Fig. \ref{day}. We noted that the test results on overcast and snow sequences are almost the same as those on sun and cloud sequences, and the difference between Recall@1 is only about 2.15\%. And the test results in rain and night sequences are poor, with Recall@1 being 14.23\% lower on average. The experimental results of model trained on all night sequences and tested on night and day sequences respectively is shown in Fig. \ref{night}. Due to the large noise of the events in the night sequences, the trained model has poor performance in the daytime scenes.

\textbf{Results on DDD17.}
In addition, in order to verify the generalization performance of our method, we use the network model trained on the synthesis event streams of Oxford RobotCar dataset, and then we test the trained model on freeway sequences of DDD17 dataset recorded by real DAVIS camera (Please refer to APPENDIX for details). Surprisingly, the Recall@5 of our method on the night freeway sequences in DDD17 dataset is up to 92.2\% without any transfer learning or fine-tuning of the trained network. It means that just using synthesis events for model training can obtain excellent results on events recorded by real event camera. All sequences results are shown as Table \ref{tab:2}.
\begin{table}[htbp]
\vspace{-2mm}
\begin{center}
\caption{Results on DDD17 of models trained with synthesis events}
\setlength{\tabcolsep}{1pt}
\newcommand{\tabincell}[2]{\begin{tabular}{@{}#1@{}}#2\end{tabular}}
\begin{tabular*}{0.95\linewidth}{@{}@{\extracolsep{\fill}}c|ccccp{1cm}@{}}
\toprule
Suquences   & Recall@1  & Recall@5  & Recall@10 & Recall@20    \\
\midrule
Night1(rec1487350455)   & 80.18\%   & 88.52\%   & 91.76\%   & 94.72\%    \\
Night2(rec1487608147)   & 79.50\%   & 92.22\%   & 95.57\%   & 97.39\%     \\
Night2(rec1487609463)   & 73.31\%   & 88.82\%   & 93.21\%   & 95.64\%     \\
Day1(rec1487417411)     & 76.80\%   & 90.43\%   & 93.65\%   & 95.97\%    \\
Day2(rec1487594667)     & 76.26\%   & 91.07\%   & 94.14\%   & 96.36\%     \\
Day3(rec1487430438)     & 78.06\%   & 91.10\%   & 94.27\%   & 96.65\%   \\
\bottomrule
\end{tabular*}
\label{tab:2}
\vspace{-6mm}
\end{center}
\end{table}

\subsection{Performance Comparison with Different VPR Methods}
In order to compare the performance of our Event-VPR with conventional frame-based method, we compared Event-VPR with NetVLAD \cite{arandjelovic2016netvlad} and SeqSLAM \cite{milford2012seqslam} on MVSEC and Oxford RobotCar datasets, and the results are shown in Table \ref{tab:3}. For fair comparison, we adjusted the spatial resolution of the image frames to be the same as that of the event bins. The results show that our method is better than NetVLAD and much better than SeqSLAM on MVSEC dataset, since most images recorded by DAVIS camera are dim and Event-VPR has advantages under this condition. On Oxford RobotCar dataset, the recall@1 of Event-VPR is about 26.02\% higher than SeqSLAM but about 7.86\% lower than NetVLAD. We think the reason is that a lot of fixed pattern noise is added to the event streams manually generated by V2E \cite{delbruck2020v2e}. Considering that the event cameras are not mature enough (e.g. low spatial resolution, high noise level, etc.), therefore Event-VPR shows similar performance to frame-based method.

\begin{table}[htbp]
\vspace{-2mm}
\begin{center}
\caption{Comparison of Event-VPR with other VPR methods on MVSEC and Oxford RobotCar Datasets}
\setlength{\tabcolsep}{1pt}
\newcommand{\tabincell}[2]{\begin{tabular}{@{}#1@{}}#2\end{tabular}}
\begin{tabular*}{\linewidth}{@{}@{\extracolsep{\fill}}c|ccccc|ccccp{1cm}@{}}
\toprule
\multirow{2}{*}{Methods} &\multicolumn{9}{c}{Recall@1 (\%)}\\ 
& day1  & day2  & night1    & night2    & night3    & cloud & rain  & snow  & night \\ 
\midrule
Event-VPR(Ours)       & \textbf{99.51}    & 91.52 & \textbf{98.67}    & \textbf{95.11} & \textbf{97.37}    & 91.81 & 90.95 & 93.29 & 91.80 \\
NetVLAD \cite{arandjelovic2016netvlad}  & 98.57 & \textbf{96.74} & 96.00     & 92.44     & 82.89     & \textbf{99.34} & \textbf{98.57} & \textbf{99.84} & \textbf{99.67} \\
SeqSLAM \cite{milford2012seqslam}   & 72.61 & 68.42 & 55.37     & 58.68     & 57.72     & 73.10    & 63.37    & 69.98    & 57.32    \\
\bottomrule
\end{tabular*}
\label{tab:3}
\vspace{-6mm}
\end{center}
\end{table}

\subsection{Impact of Different Event Representations}
\begin{figure}[htbp]
\centering
\vspace{-1mm}
\includegraphics[width=0.95\columnwidth]{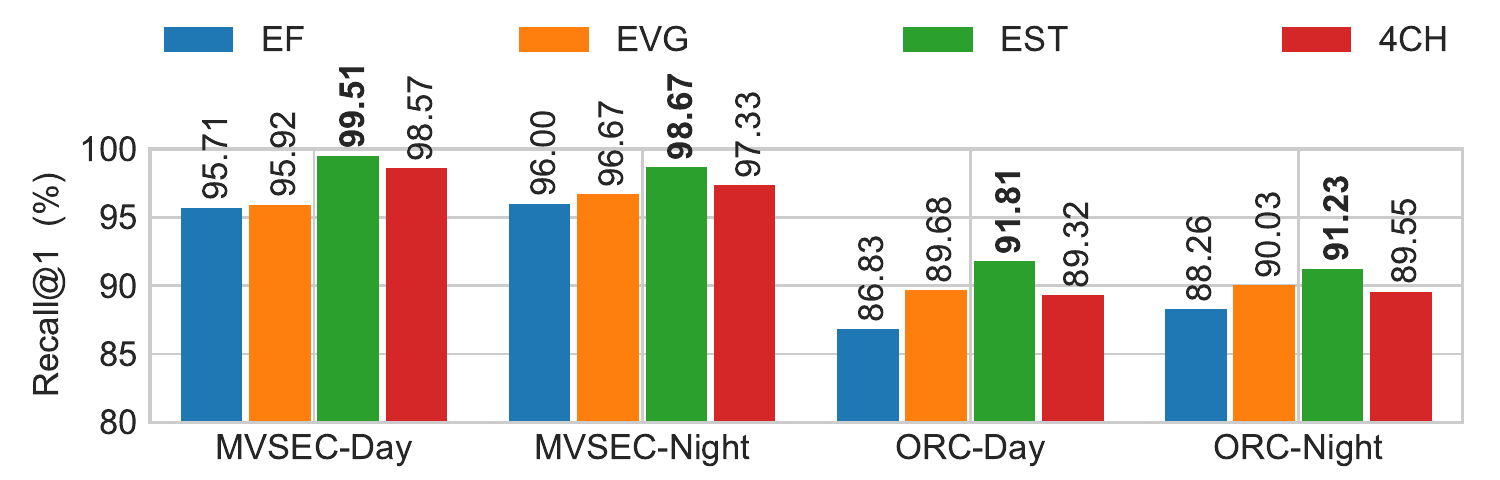}
\caption{Comparison of the Performance Impact of Different Event Representation about Event-VPR (MVSEC \& Oxford RobotCar Datasets).}
\label{fig:7}
\vspace{-5mm}
\end{figure}

In this experiment, we explored the Event-VPR performance with different event representations. In addition to the EST voxel grid described in our method, we have also tried several other event representations, including event frame (EF), unipolar event voxel grid (EVG), and 4-channel event count and last-timestamp image (4CH). We all used ResNet34 as the feature extraction network in this experiment. As shown in Fig. \ref{fig:7}, since EF discards timestamps and EVG discards event polarity, EST voxel grid has great advantages over other representations. Due to 4CH discards earlier timestamps, it is also not as good as EST voxel grid. EST voxel grid not only retains the local spatio-temporal neighborhood information of events, but also extracts effective features through using learnable kernel to suppress the noise of events. 

\subsection{Impact of Different Network Structures}
\begin{figure}[htbp]
\centering
\vspace{-1mm}
\includegraphics[width=0.95\columnwidth]{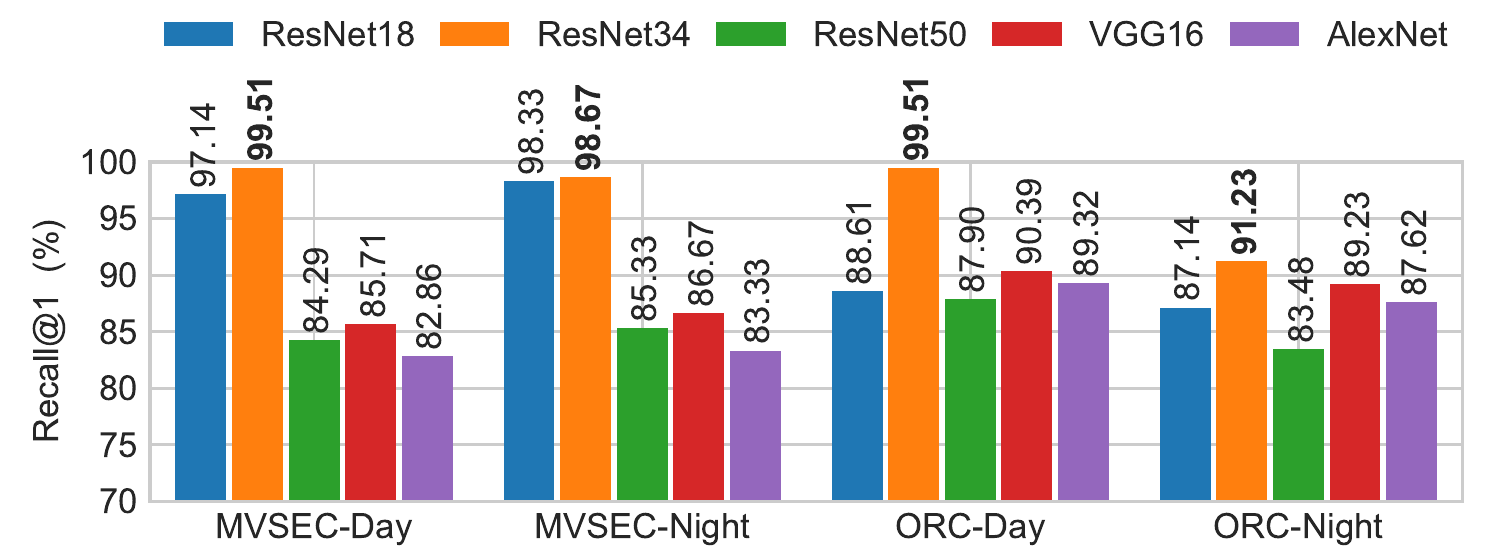}
\caption{Comparison of the Performance Impact of Different Feature Extraction Networks about Event-VPR (MVSEC \& Oxford RobotCar Datasets).}
\label{fig:8}
\vspace{-3mm}
\end{figure}

In this experiment, we explored the Event-VPR performance of different network structures. In addition to the ResNet34 described in our method, we also trained VGG-16, AlexNet and two other ResNet with different network capacities. We used EST voxel grid as the event representation in the whole experiment. As shown in Fig. \ref{fig:8}, experimental results show that different feature extraction networks have no significant impact on the network performance, while a larger network increases the computing resources without improving the accuracy. We guess that a larger network needs more parameters, which makes the network easier to overfitting. We finally choose ResNet34 as the feature extraction network of our pipeline. However, we can replace it with other different convolutional networks according to our own needs and computing resources.

\section{CONCLUSIONS}
\label{sec:conclusions}
We proposed an end-to-end method (Event-VPR) to solve the problem of large-scale place recognition by using event cameras. We showed that our Event-VPR method based on events is more robust to changes in the surrounding environment caused by weather and time changes than conventional frame-based VPR methods. Though we get very good results, it should be noted that the spatial resolution of event cameras is still very low, so there are still some deficiencies compared to frame-based VPR methods. In the future, we will try to combine standard cameras and event cameras to implement a hybrid network architecture for visual place recognition based on frames and events.






\bibliographystyle{IEEEtran}
\bibliography{mybibfile}

\section*{APPENDIX}
\label{sec:appendix}

\subsection{Details on Network Training}

\textbf{Caching.}
In the process of network training, it is necessary to retrieve the descriptor vector of samples to calculate the descriptor distance. In order to improve the training efficiency, we build a cache for the descriptor vectors of the entire database and use the cached descriptor vectors to select the best positive and hard negatives. Because the model parameters will be continuously adjusted during the training process and the descriptor vectors of the network output will also be continuously changed, we will update the cache regularly during the network training. Recalculating the cache every 500-1000 training queries can achieve a good balance between the duration of the training period, the convergence speed of the network and the quality of the model.

\textbf{Training Data Loading.}
Different from traditional image frames, sparse event bins have the problem that the amount of data is not fixed in the process of loading data, which makes it difficult to distinguish each sample during batch training.
When we build a batch process, we add a column of indexes to each event bins, arrange the query, the best positive and the hard negatives in turn, merge them into a batch process and send them to the network for training. Then the descriptor vectors are divided according to the indexes.

\subsection{Dataset Configuration}
In our experiments, we use MVSEC \cite{zhu2018multivehicle}, DDD17 \cite{binas2017ddd17} and Oxford RobotCar datasets \cite{maddern20171}. 
The MVSEC and DDD17 datasets are event-based driving datasets recorded in the real environment. We select intensity images and event bins of 5 outdoor driving sequences from MVSEC dataset (including day-and-night scenes, using left-eye DAVIS camera) and 6 outdoor driving sequences from DDD17 dataset (including urban and freeway scenes), which include dramatic illumination changes and scene structure changes. In addition, Oxford RobotCar dataset is a standard dataset commonly used in VPR. We use the event synthesizer V2E \cite{delbruck2020v2e} to convert the image sequences recorded by the center camera of the three-eye vision sensor (Bumblebee XB3) in Oxford RobotCar dataset into corresponding event streams. We tried our best to select sequences with the same trajectory under different weather conditions, including sun, cloud, rain, snow and night scenes. We think that such scenarios are more challenging for testing the performance of VPR methods.

We randomly divide the sequences of the same trajectory into geographically non-overlapping training and test sets (detailed results are shown in Table \ref{tab:4} \ref{tab:5} \ref{tab:6} \ref{tab:7} \ref{tab:8}). In the MVSEC dataset, we select about 40k training samples and 10k test samples from 5 sequences. In the DDD17 dataset, we select about 120k test samples from 6 sequences. In Oxford RobotCar dataset, we also select about 50k training samples and 12k test samples from 11 sequences. For parameters, we choose potential positive distance threshold $\lambda=10\mathrm{m}$, potential negative distance threshold $\delta=25\mathrm{m}$, and true positive geographical distance threshold $\varphi=20\mathrm{m}$ in MVSEC and Oxford Robocar datasets. Moreover, all parameters in DDD17 dataset are the same except that the true positive geographical distance threshold is $\varphi=60\mathrm{m}$.

Generally, the performance of VPR methods is evaluated by Recall. In this paper, we use Recall@N to evaluate the performance of our algorithm. Specifically, for each query, the N-nearest neighbor samples are retrieved as positive samples according to the descriptor distance. If at least one of samples is less than $\varphi$ away from the query according to their geographical distance, it is considered to be correctly identified. Then, we calculate the Recall@N with different N. For the query set, we traverse descriptor vectors of all queries to calculate Recall@N which is the percentage of correctly identified query descriptors. In our experiments, we restrict our analysis to $\mathrm{N}=\{1,5,10,20\}$.

\begin{figure}[h]
\centering
\includegraphics[width=0.95\columnwidth]{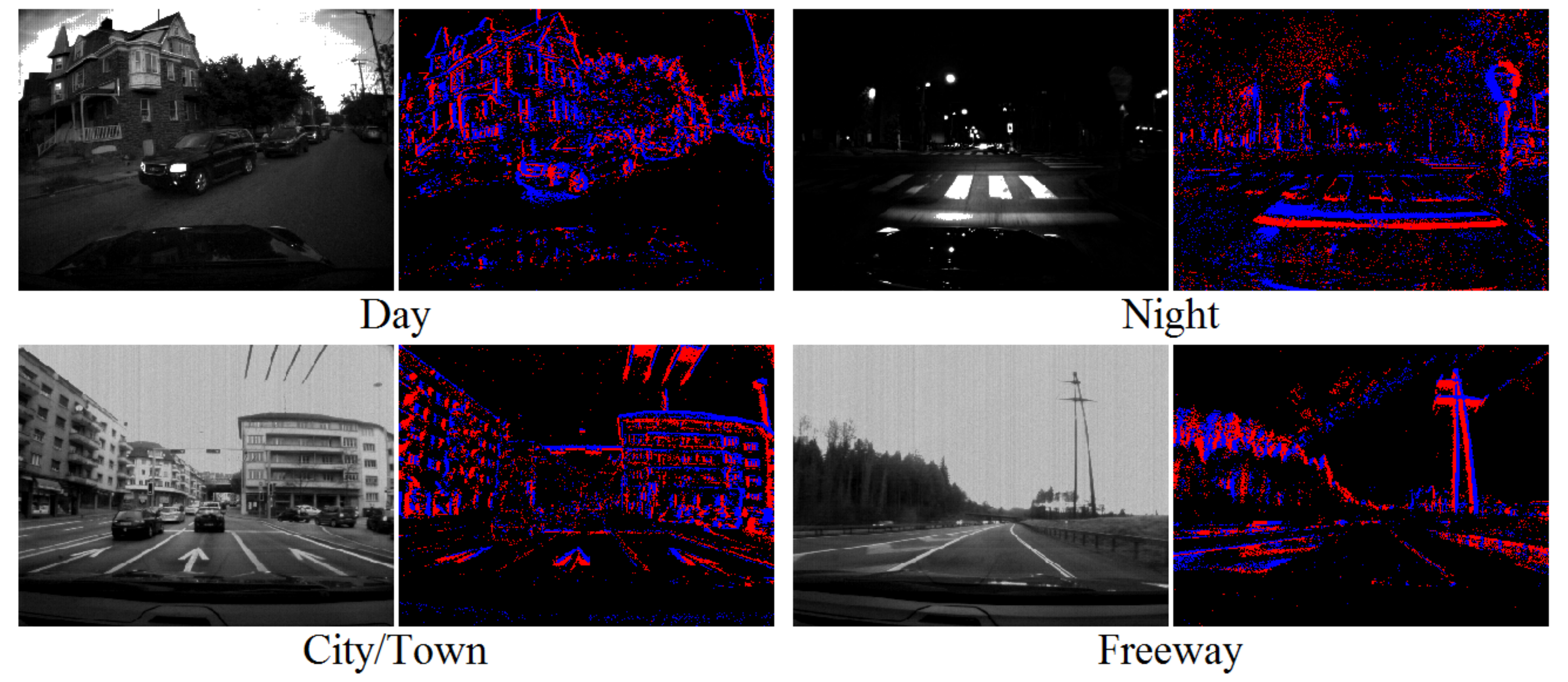}
\caption{The Scenes of MVSEC and DDD17 Datasets in Our Experiments.}
\label{fig:9}
\vspace{-3mm}
\end{figure}

\begin{table}[ht]
\begin{center}
\caption{Typical Values and Parameter Choices in the MVSEC Dataset}
\setlength{\tabcolsep}{1pt}
\newcommand{\tabincell}[2]{\begin{tabular}{@{}#1@{}}#2\end{tabular}}
\begin{tabular}{l|cccccp{1cm}}
\toprule
Parameters                          & Day1  & Day2  & Night1& Night2& Night3\\
\midrule
Total Data $N$                      & 11937 & 28583 & 2704  & 3863  & 2854  \\
Training Database $N_{db,train}$    & 7140  & 17199 & 1597  & 2274  & 1678  \\
Training Database $N_{q,train}$     & 2428  & 5760  & 577   & 752   & 620   \\
Test Database $N_{db,test}$  	    & 1804  & 4303  & 412   & 590   & 445   \\
Test Database $N_{q,test}$          & 615   & 1440  & 151   & 226   & 153   \\
Random Negatives $N_{n,r}$	        & 300   & 400   & 100   & 100   & 100   \\
Hard Negatives $N_{n,h}$            & 10    & 10    & 10    & 10    & 10    \\
\bottomrule
\end{tabular}
\label{tab:4}
\vspace{-5mm}
\end{center}
\end{table}

\begin{table}[ht]
\begin{center}
\caption{The Sequences of the DDD17 Dataset used in Our Experiments}
\setlength{\tabcolsep}{1pt}
\newcommand{\tabincell}[2]{\begin{tabular}{@{}#1@{}}#2\end{tabular}}
\begin{tabular}{l|cp{1cm}}
\toprule
Freeway (Day)   & rec1487417411 / rec1487419513 / rec1487430438 \\
\midrule
Freeway (Night) & rec1487350455 / rec1487608147 / rec1487609463 \\
\bottomrule
\end{tabular}
\label{tab:5}
\vspace{-5mm}
\end{center}
\end{table}

\begin{table}[ht]
\begin{center}
\caption{Typical Values and Parameter Choices in the DDD17 Dataset}
\setlength{\tabcolsep}{1pt}
\newcommand{\tabincell}[2]{\begin{tabular}{@{}#1@{}}#2\end{tabular}}
\begin{tabular}{l|cccccp{1cm}}
\toprule
Parameters                  & Day1  & Day2  & Day3  & Night1 & Night2 & Night3  \\
\midrule
Total Data $N$              & 20949 & 29846 & 31336 & 14023  & 12072  & 14571   \\
Test Database $N_{db,test}$ & 16759 & 23876 & 25068 & 11218  & 9657   & 11656   \\
Test Database $N_{q,test}$  & 4190  & 5970  & 6268  & 2805   & 2415   & 2915    \\
Random Negatives $N_{n,r}$  & 500   & 500   & 500   & 500    & 500    & 500     \\
Hard Negatives $N_{n,h}$    & 10    & 10    & 10    & 10     & 10     & 10      \\
\bottomrule
\end{tabular}
\label{tab:6}
\vspace{-5mm}
\end{center}
\end{table}

\begin{table}[ht]
\begin{center}
\caption{The Sequences of the Oxford RobotCar Dataset used in Our Experiments}
\setlength{\tabcolsep}{1pt}
\newcommand{\tabincell}[2]{\begin{tabular}{@{}#1@{}}#2\end{tabular}}
\begin{tabular}{l|cp{1cm}}
\toprule
Sun/Cloud   & 2014-11-18-13-20-12 / 2015-03-10-14-18-10   \\
            & 2015-07-29-13-09-26 / 2015-09-02-10-37-32   \\
\midrule
Overcast    & 2015-02-13-09-16-26 / 2015-05-19-14-06-38   \\
\midrule
Rain        & 2014-11-25-09-18-32 / 2014-12-05-11-09-10   \\
\midrule
Snow        & 2015-02-03-08-45-10   \\
\midrule
Night       & 2014-11-14-16-34-33 / 2014-12-10-18-10-50   \\
\bottomrule
\end{tabular}
\label{tab:7}
\vspace{-5mm}
\end{center}
\end{table}

\begin{table}[ht]
\begin{center}
\caption{Typical Values and Parameter Choices in the Oxford RobotCar Dataset}
\setlength{\tabcolsep}{1pt}
\newcommand{\tabincell}[2]{\begin{tabular}{@{}#1@{}}#2\end{tabular}}
\begin{tabular}{l|cccccp{1cm}}
\toprule
Parameters                  & Sun/Cloud & Overcast  & Rain  & Snow  & Night \\
\midrule
Total Data $N$                      & 23539 & 12002 & 10909 & 6001  & 12002 \\
Training Database $N_{db,train}$    & 13284 & 7291  & 6627  & 3578  & 7125  \\
Training Database $N_{q,train}$     & 4384  & 2356  & 2142  & 1218  & 2451  \\
Test Database $N_{db,test}$  	    & 3268  & 1797  & 1634  & 908   & 1810  \\
Test Database $N_{q,test}$          & 1212  & 558   & 507   & 298   & 622   \\
Random Negatives $N_{n,r}$	        & 500   & 500   & 500   & 500   & 500   \\
Hard Negatives $N_{n,h}$            & 10    & 10    & 10    & 10    & 10    \\
\bottomrule
\end{tabular}
\label{tab:8}
\vspace{-5mm}
\end{center}
\end{table}



\end{document}